\pdfoutput=1

\documentclass[11pt]{article}

\usepackage[]{ACL2023}

\usepackage{times}
\usepackage{latexsym}
\usepackage{graphicx}

\usepackage{colortbl}
\usepackage{subcaption}
\usepackage[export]{adjustbox}
\usepackage{listings}

\usepackage{booktabs}
\usepackage{multirow}

\usepackage[T1]{fontenc}

\usepackage[utf8]{inputenc}

\usepackage{microtype}

\usepackage{inconsolata}

\title{Toward Reliable Ad-hoc Scientific Information Extraction: A Case Study on Two Materials Datasets}

\author{Satanu Ghosh$^{1}$ \quad Neal R. Brodnik$^{2}$ \quad Carolina Frey$^{2}$\quad Collin Holgate$^{2}$ \\
\bf Tresa M. Pollock$^{2}$ \quad \bf Samantha Daly$^{2}$ \quad \bf Samuel Carton$^{1}$  \\
        $^{1}$University of New Hampshire \\ 
        $^{2}$University of California, Santa Barbara \\
         \texttt{\{satanu.ghosh, samuel.carton\}@unh.edu} \\
         \texttt{\{nrbodnik, cfrey, holgate, tresap, samdaly\}@ucsb.edu} \\}

\begin{document}
\maketitle
\begin{abstract}
We explore the ability of GPT-4 to perform ad-hoc schema based information extraction from scientific literature. We assess specifically whether it can, with a basic prompting approach, replicate two existing material science datasets, given the manuscripts from which they were originally manually extracted. We employ materials scientists to perform a detailed manual error analysis to assess where the model struggles to faithfully extract the desired information, and draw on their insights to suggest research directions to address this broadly important task.
\end{abstract}

\section{Introduction}

\begin{figure}[h!]
    \centering
    \includegraphics[width=0.5\textwidth]{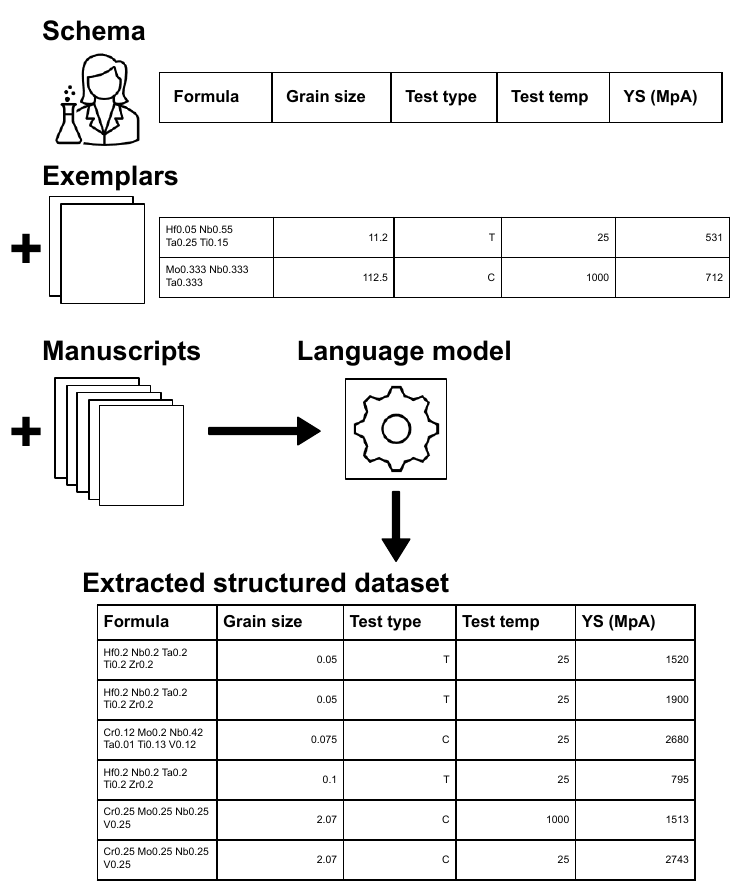}
    \caption{Extracting large-scale structured data from scientific literature should be as simple as specifying a schema, a corpus of manuscripts, and a few exemplars, and letting the LLM perform the extraction.}
    \label{fig:goal_diagram}
 \end{figure}

A key use case for large language models in science is \textbf{ad-hoc schema-based information extraction}. In this scenario, a scientist has a specific information need, such as the compressive yield strength of known multi-principal element alloys (MPEAs), and wants to (1) identify papers containing relevant information instances, and (2) comprehensively extract these instances according to a desired schema, which might include secondary information such as testing conditions and synthesis details (Fig \ref{fig:goal_diagram}). 

Such datasets are a key element of scientific informatics, transforming information reported inconsistently throughout the literature into a structured format appropriate for use in machine learning. In materials science, such datasets are used in a wide range of applications, including model-driven materials discovery, where they can be used to train or warm-start surrogate models of material properties which are then incorporated into various kinds of discovery algorithms \cite{pilania_machine_2021}.


This is a challenging extraction task. The materials science literature contains a plethora of experimental results, often spanning multiple decades, in a multitude of different formats with varying physical units. Prior generations of models needed large amounts of data to fine-tune, and were brittle and non-transferable, limiting their practicality for this type of ad-hoc, one-off extraction task. To date, manual annotation is still the standard approach for extraction of this type of unstructured data \cite{xu_small_2023}.  

Contemporary large language models such as GPT-4 \cite{openai_gpt-4_2023} and Gemini \cite{team_gemini_2023} have the potential to conquer this barrier. With the emergent few-shot capabilities of these models, a scientist can potentially present a LLM with a (1) collection of manuscripts, (2) a schema defining what types of data they want to extract, and (3) a few exemplars of performing this extraction, and receive back a tabular dataset of information extracted from those manuscripts, ready to use for their chosen purpose.  This ad-hoc paradigm  has been referred to as ``on-demand'' information extraction (ODIE) \cite{jiao_instruct_2023} and it promises to speed up the search for new materials with unique properties. 

We argue that this very ambitious extraction setting should be a key target for LLM-driven scientific information extraction, as it can potentially allow scientists to extract a maximum of clinically relevant data with a minimum of effort, both in materials science and other domains such as biomedicine. Two significant questions, then, are \textbf{(1) to what extent can contemporary LLMs perform this type of ad-hoc extraction on scientific text, and (2) what are the major barriers impeding their effectiveness?}

In this paper, we explore these questions using two manually-extracted material property datasets, one focused on multi-principal element alloys (MPEAs), sometimes known as high-entropy alloys (HEAs) \cite{borg_expanded_2020}, and one focused on elemental diffusion in silicate melts \cite{zhang_diffusion_2010}. We use these existing datasets as sources of schema structure and exemplars, and evaluate whether a representative contemporary LLM (GPT-4) is capable of replicating them given the manuscripts from which they were originally extracted. We perform a detailed error analysis, including annotation by domain experts, to understand where and why the model falls short, to guide further development of LLMs to advance extraction paradigms for materials data.


Generally, we find that the model shows great potential for extracting information described narratively or in the form of a conventionally-formatted table. The majority of errors are attributable to figures, which our current pipeline does not address, and PDF parsing issues.  Other sources of error include non-standard table formats, the need for additional postprocessing of extracted values, as well as true reading comprehension errors which could be addressed via improved prompt engineering. All relevant code and data can be found in our github repository \footnote{\url{https://github.com/SatanuG/ad_hoc_information_extraction}}.

\section{Related work}

Information extraction (IE) is a core NLP task with a long history, and one of many which contemporary large language models (LLMs) such as GPT-4 \cite{openai_gpt-4_2023} 
and Gemini \cite{team_gemini_2023} 
are able to perform in a zero- or few-shot setting \cite{xu_large_2023}. 
This few-shot extraction ability has enabled a new extraction paradigm, involving extracting a complex schema with little or no training data, termed ``on-demand information extraction'' (ODIE) by \cite{jiao_instruct_2023}. Our task setting falls into this paradigm. 

Zero- and few-shot scientific information extraction forms a major aspect of the excitement over LLMs' potential for  accelerating scientific discovery, because of its potential to create structured information from diffuse unstructured data sources such as scientific papers and clinical notes \cite{hope_computational_2022,morris_scientists_2023}. Ad-hoc extraction from scientific literature is particularly challenging because the amount of manually-annotated data needed to fine-tune a model to perform a particular extraction task would constitute a significant fraction of all the data in existence, obviating the benefit of training such a model. Scientific literature is also highly multimodal, with key information presented in narrative, tabular, and visual formats. SciDaSynth \cite{wang_scidasynth_2024} approaches the problem with an interactive QA-based interface, evaluating based on a 12-scientist user study.

 LLM-powered IE has attracted attention in materials science specifically \cite{olivetti_data-driven_2020, xie_large_2023, polak2024extracting}. \citet{polak_flexible_2024} explore the use of GPT-4 to extract bulk modulus information from sentences extracted from scientific papers, while \citet{yang_accurate_2023} uses zero-shot ChatGPT with a verification step to extract band gap information from sentences collected in a dataset by \cite{dong_auto-generated_2022}. \citet{walker_extracting_2023} fine-tunes GPT-3.5 to extract growth procedures and outcomes for gold nanorods from paper text. \citet{montanelli_high-throughput_2024} uses the Cohere model to extract phase-property relationships from a corpus of full-text materials papers. Finally, the A-Lab \cite{szymanski_autonomous_2023} incorporates extracted synthesis recipes into an AI-guided experimentation protocol. Our work differs from these recent approaches in seeking to extract a rich, complex schema from non-curated data in one pass with no fine-tuning, instead of narrowly-defined tasks over carefully-curated corpora. This is a highly ambitious and difficult extraction task, but one that places a minimum of burden on the scientist, and we see it as a key milestone for scientific IE. 

\begin{table}[]
\begin{tabular}{@{}llll@{}}
                           &           & \textbf{MPEA} & \textbf{Diffusion} \\ \midrule
\multirow{2}{*}{Total}  & \# Papers & 164           &   71             \\
                           & \# Rows   & 1211          &   2838            \\ \midrule
\multirow{2}{*}{Retrieved} & \# Papers & 128          & 55               \\
                           & \# Rows   & 971          & 2359               \\ \midrule
\multicolumn{2}{l}{Publication year}   & 2004-2022     & 1964-2009          \\ \bottomrule
\end{tabular}

\caption{Basic statistics for the two datasets, publication year range. For both datasets, a subset of total papers were retrieved, parsed and analyzed in this work.}
\label{tab:dataset_stats}
\end{table}

\section{Datasets}

We experiment with two material properties datasets, one pertaining to multi-principal element alloys (MPEAs) \cite{borg_expanded_2020}, and one to elemental diffusion in silicate melts \cite{zhang_diffusion_2010}. Table \ref{tab:dataset_stats} reports basic statistics of each dataset, including the number of unique papers, number of unique records (since a single paper can yield multiple records), and publication year range.  Table \ref{tab:dataset_column_descriptions} describes the individual columns in each dataset.

\paragraph{MPEA Mechanical Properties Dataset}


Multi-principal element alloys (MPEAs) are a new class of chemically complex metallic alloys of interest for applications in energy, defense, and transportation, distinguished from conventional alloys by having relatively even proportions of many elements \cite{miracle_critical_2017,miracle_refractory_2020}. The MPEA dataset used for this analysis specifically aims to record key properties reported for MPEAs derived primarily from the refractory metals (Cr, Hf, Mo, Nb, Ru, Ta, Ti V, W, Zr). 
These properties include yield strength, which represents the force necessary to make a metal permanently deform, and elongation, which represents the degree to which a material can deform prior to failure.
Hardness, which represents the ability of the material resist localized deformation, can be used to estimate the yield strength of a material.
Other useful details recorded in the MPEA dataset where available include the concentration of interstitial elements present (i.e. oxygen, nitrogen, and carbon concentrations), the method by which the material was processed 
, the test temperature, and the method of testing.

The MPEA dataset used in this study is an expansion of the dataset published in \cite{borg_expanded_2020}, which has been continually updated . Records were taken from experimental literature published in 2004-2022 and are identified by a Digital Object Identifier (doi) that was assigned by the publisher.

\paragraph{Diffusion in Silicate Melts Dataset}
Our second dataset \cite{zhang_diffusion_2010} covers the diffusion (transport of elements) inside silicate glasses and melts (e.g., magma), which affects many processes in geology and materials science. The key value in this dataset is \textit{diffusion coefficient}, which is often  symbolized as $D$ and has units of m\textsuperscript{2}/s. The diffusion coefficient is a proportionality constant that describes how rapidly one element is transported within a substance. As one may expect, the value of $D$ depends on what the diffusing element is, the medium (magma) it is moving in (i.e., its composition), and also environmental conditions like temperature, pressure, and how much water is present (which greatly affects magmatic structure). The diversity of magmas found in the earth and the 
elements being transported within them motivates building and maintaining databases of diffusion coefficients to help elucidate underlying truths.

Outside of its fundamental value, we chose this dataset for our study because (1) diffusion as a process is heavily sensitive to experimental conditions, meaning value extraction is only useful with the associated large quantity of metadata; (2) it is a large dataset, affording sufficient measurements and papers to draw meaningful conclusions; (3) the multi-decade span of publication dates offers exposure to a range of scientific conventions and pdf quality levels from the perspective of typesetting, character recognition, and table format; and (4) it represents a field where updates occur on a decadal basis, thus valuing a tool for automatic extraction of new measurements.


\section{Method}

Our extraction pipeline consists of the following steps: (1) retrieving and parsing source PDFs; (2) prompting and response postprocessing; and (3) extracted row alignment. 



\subsection{Retrieving and parsing source PDFs}
The vast majority of PDFs available in both datasets are not open-access, and are referred to only by their DOI within each dataset. As a first stage, we manually download as many PDFs as we can easily access, resulting in 128 PDFs of the original 164 for the MPEA dataset, and 55 out of 71 for the diffusion dataset. Our github repository includes the DOIs of the PDFs we downloaded but not, for copyright reasons, the PDFs of these papers. 


As of June 2024, none of the major LLM APIs allow programmatic uploading of PDF files. Therefore, to process PDFs at scale, they must be converted to an intermediate readable format. We use GROBID\footnote{\url{https://github.com/kermitt2/grobid}} to convert paper PDFs to XMLs. When parsed correctly, the  XML contains all textual information from the paper, including tables and their respective captions. However, GROBID cannot process figures, leaving any figure-bound information inaccessible by definition. The error analysis presented in Section \ref{subsec:error_analysis} assesses the magnitude of this problem. 


\subsection{Prompting}

We experiment with three basic prompting techniques: zero- and one-shot prompting, and LangChain. For each technique, we try both a ``complex'' schema including all columns from the original dataset, and a reduced ``simple'' schema including only a subset of column. 

\begin{figure}[h!]
    \centering
    \includegraphics[width=0.5\textwidth]{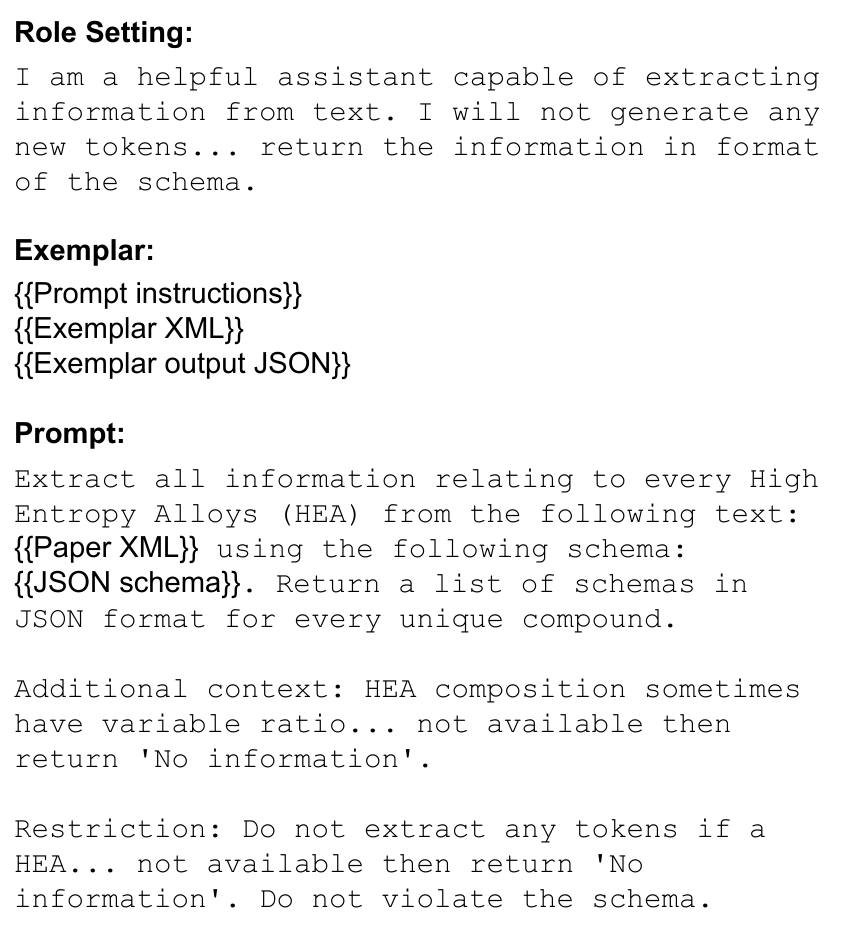}
    \caption{Abbreviated one-shot prompt. The prompt begins with role-setting, includes a single exemplar with prompt instructions, then repeated prompt instructions with additional context and clarifications.}
    \label{fig:prompt_schematic}
 \end{figure}

\paragraph{Zero- and one-shot prompting}

Working with the \texttt{gpt-4-1106-preview} GPT-4 Turbo model variant via the OpenAI API, we use a very basic and general prompting approach, illustrated in Fig. \ref{fig:prompt_schematic}.
Our goal is to extract all columns from a single full paper XML at once, requesting them in the form of a JSON. In the one-shot condition, the prompt includes a single exemplar, selected randomly to be \cite{long2019fine} and \cite{russel2004effect} for the MPEA and diffusion dataset respectively. 

We additionally make use both role-setting and task instructions in our prompt. Role-setting has been found to be generally helpful in prompt engineering \cite{shanahan_role_2023, kong_better_2023}, so we set various desired aspects of the LLM's behavior via this mechanism, i.e., extracting information from the given text without generating new tokens and following the provided schema (if any). The task instructions are specific and detailed, customized to each dataset based on expert input and include a schema that should be followed. We add additional context with background knowledge and special instructions. A more complete version of the prompt is available in the appendix (Section \ref{sec:full_prompt_details}. 

\paragraph{LangChain} LangChain\footnote{\url{https://python.langchain.com/v0.2/docs/introduction/}} is a popular open-source framework for developing solutions that combine LLMs with other tools. Information extraction is one of the many use cases that can be performed with LangChain. We follow the guidelines on their website to perform structured information extraction for long text. In short, the LLM is provided with a schema of various entities and a short description of each entity. In our use case, these entities are the column headers in the two datases.  An important distinction of this method from basic prompting methods was \textit{chunking the papers into smaller blocks of 2000 tokens}, with each block having a 1\% overlap with the previous one.

\paragraph{Postprocessing}
Minimal postprocessing is needed for the model's JSON output, which varies slightly in format from run to run. This includes wrapping JSON-formatted output in `\texttt{`\hspace{0cm}`\hspace{0cm}`json}'  tags. The model also sometimes produces Javascript-style `\texttt{//}' comments explaining its extraction decisions; therefore, we use the \texttt{jstyleson} Python library\footnote{\url{https://github.com/linjackson78/jstyleson}} to parse this JSON output rather than the default Python \texttt{json} library. 

\subsection{Extracted row alignment}
\label{para:alignment}
In order to evaluate our success in replicating the original datasets, including which rows were matched, which original rows were missed, and which extracted rows were hallucinated, we first need to determine which extracted row corresponds with which original dataset row. This is nontrivial, because the columns that define a unique record will vary from paper to paper. One paper might test multiple different compositions, while another might test the same composition under different testing conditions, while a third might perform multiple trials of the same experiment. Given that the model will miss or hallucinate many individual cell values, this becomes a nuanced matching task. 

Our approach is to combine domain-specific minimum criteria for a possible match with a greedy matching process within each paper. An extracted row is a potential match for an original row if it shares values for hand-chosen required columns, which we chose to be \textbf{formula and yield strength} for the MPEA dataset and \textbf{diffusing element and diffusion coefficient} for the diffusion dataset. Beyond this minimum, match strength is determined by proportion of shared column values between the two rows, after which rows are greedily paired with each other until no possible pairs remain for a given paper. Paired rows are classified as a `match', unpaired rows from the original dataset are classified as a `miss', and unpaired extracted rows are classified as a `hallucination'.


\section{Automated evaluation}


\begin{table}[]
\resizebox{\columnwidth}{!}{%
\begin{tabular}{llllll}
\hline
\textbf{Dataset}                          & \textbf{Method} & \textbf{Matched} & \textbf{Recall} & \textbf{Hallucinated} & \textbf{Precision} \\ \hline
\multicolumn{1}{c}{\multirow{6}{*}{MPEA}} & 0-shot-simple   & 56               & 0.062           & 166                   & 0.245              \\
\multicolumn{1}{c}{}       & 0-shot-complex    & 56  & 0.066 & 235  & 0.369 \\
\multicolumn{1}{c}{}       & 1-shot-simple     & 243 & 0.280 & 259  & 0.502 \\
\multicolumn{1}{c}{}       & 1-shot-complex    & 174 & 0.202 & 207  & 0.475 \\
\multicolumn{1}{c}{}       & langchain-simple  & 188 & 0.253 & 746  & 0.219 \\
\multicolumn{1}{c}{}       & langchain-complex & 129 & 0.182 & 564  & 0.213 \\ \hline
\multirow{6}{*}{Diffusion} & 0-shot-simple     & 0   & 0     & 93   & 0     \\
                           & 0-shot-complex    & 0   & 0     & 126  & 0     \\
                           & 1-shot-simple     & 2   & 0.001 & 52   & 0.037 \\
                           & 1-shot-complex    & 23  & 0.013 & 193  & 0.106 \\
                           & langchain-simple  & 70  & 0.032 & 1606 & 0.060 \\
                           & langchain-complex & 47  & 0.022 & 1423 & 0.048 \\ \hline
\end{tabular}%
}
\caption{Extraction performance based on match, missed, and hallucinated values on different variations of dataset and methods.}
\label{tab:result_basic}
\end{table}

After aligning extracted rows with dataset rows as described above, we report the number of matches and hallucinations in Table \ref{tab:result_basic} along with the recall and precision. For both the datasets, two variations of schema and three types of prompting. The only difference between simple and complex schema is the number of properties mentioned in each schema.  

The raw performance is underwhelming. On MPEA, the model hallucinates roughly as many rows as it matches, and misses twice as many. We are able to match significantly more entries using the simple schema than the complex schema. While using Langchain we extracted more relevant entries, but we can see a drop of precision (more hallucination). On the Diffusion dataset, we are able to match no rows at all for 0-shot and, though the model hallucinates at a proportionally lower rate than for the MPEA dataset. Using Langchain clearly improved the performance on this dataset but the hallucinations also increased significantly. 

Table \ref{tab:results_table} gives a detailed breakdown of extraction performance for related properties (all properties other than the ones used to match the entries), properties of high variance (experimental properties that were subjected to change in a paper) and properties of low variance (experimental constants). 
The two different schemas present the two different notions of information extraction: maximal and minimal. Tasks that require less exploration of contextually related information extraction can follow the minimal approach. 
From the results, we can see that the extraction of fewer properties aids the model, as it allows more useful extractions with higher recall (on the primary property). The precision value gives us an idea about the hallucinated extractions, and the simpler schema resulted in lesser hallucination compared to the complex schema. The 0-shot prompting did not perform well for low-variance properties, but performed much better at extracting high-variance properties. Langchain performs better than 0-shot in most cases but outperformed by the 1-shot, considering that we cannot draw any conclusion from the Diffusion dataset because the number of extractions are so low.

\begin{table*}[]
\resizebox{\textwidth}{!}{%
\begin{tabular}{@{}cccccccccccccc@{}}
\toprule
 &
   &
  \multicolumn{4}{c}{\textbf{0-shot}} &
  \multicolumn{4}{c}{\textbf{1-shot}} &
  \multicolumn{4}{c}{\textbf{Langchain}} \\ \midrule
 &
   &
  \multicolumn{2}{c}{simple} &
  \multicolumn{2}{c}{complex} &
  \multicolumn{2}{c}{simple} &
  \multicolumn{2}{c}{complex} &
  \multicolumn{2}{c}{simple} &
  \multicolumn{2}{c}{complex} \\ \midrule
 &
   &
  \textit{Recall} &
  \textit{Precision} &
  \textit{Recall} &
  \textit{Precision} &
  \textit{Recall} &
  \textit{Precision} &
  \textit{Recall} &
  \textit{Precision} &
  \textit{Recall} &
  \textit{Precision} &
  \textit{Recall} &
  \textit{Precision} \\ \midrule
 &
 
  related properties &
  0.360 &
  0.245 &
  0.311 &
  0.163 &
  0.323 &
  0.248 &
  0.269 &
  0.180 &
  0.449 &
  0.254 &
  0.301 &
  0.155 \\
 &
  low-variance properties &
  0.071 &
  0.082 &
  0.146 &
  0.146 &
  0.457 &
  0.555 &
  0.289 &
  0.301 &
  0.287 &
  0.287 &
  0.076 &
  0.076 \\
\multirow{-4}{*}{\textbf{MPEA}} &
  high-variance properties &
  0.408 &
  0.273 &
  0.394 &
  0.171 &
  0.301 &
  0.197 &
  0.259 &
  0.119 &
  0.472 &
  0.249 &
  0.391 &
  0.187 \\ \midrule
 &
  
  related properties &
  no match &
  no match &
  no match &
  no match &
  0.500 &
  0.500 &
  0.147 &
  0.138 &
  0.333 &
  0.333 &
  0.125 &
  0.333 \\
 &
  low-variance properties &
  no match &
  no match &
  no match &
  no match &
  0.500 &
  0.500 &
  0.250 &
  0.250 &
  0.500 &
  0.500 &
  0.250 &
  0.250 \\
\multirow{-4}{*}{\textbf{Diffusion}} &
  high-variance properties &
  no match &
  no match &
  no match &
  no match &
  0.500 &
  0.500 &
  0.112 &
  0.104 &
  0.250 &
  0.250 &
  0.083 &
  0.500 \\ \bottomrule
\end{tabular}%
}
\caption{Extraction results for both the datasets can be found in this table. Recall and precision are calculated based on the matched entries found in Table. \ref{tab:result_basic} (if no match is found in required properties, then we cannot report any other results).}
\label{tab:results_table}
\end{table*}

\section{Manual error analysis}

\begin{figure}[h!]
   \centering
    \subfloat[MPEA dataset]{%
        \includegraphics[clip,width=.75\columnwidth]{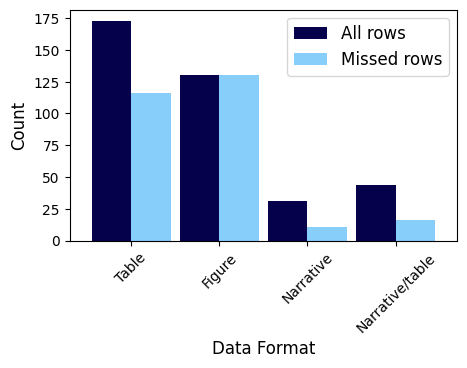}%
    }
    \vspace{1pt}
    \subfloat[Diffusion dataset]{
        \includegraphics[clip,width=.75\columnwidth]{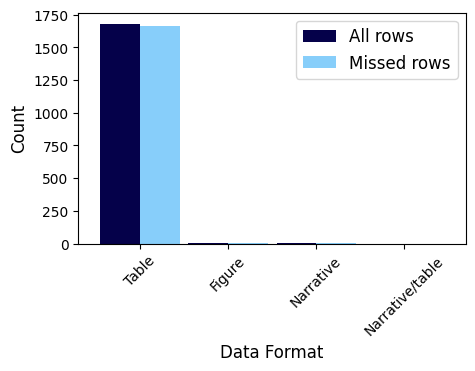}
    } 
    
    \caption{Counts of rows with key information in different formats.}
    \label{fig:data_info_err_distribution}
   
 \end{figure}

To elucidate what formats key information occurs in, and under what circumstances the model struggles to extract that information, two materials science experts hand-annotated a sample of papers and attempted LLM extractions from each dataset, using the 1-shot prompting variant. 60 papers' worth of attempted extractions were annotated from the MPEA dataset, constituting a total of 428 rows, and 40 papers from the Diffusion dataset, constituting 1714 rows. Both sets were sampled randomly from the set of retrieved and parsed papers for which extractions were attempted. Each row was labeled for the format in which the key information occurred (table, figure, narrative text, etc). Each failure, either missing or hallucinated rows, was annotated for its failure reason, to the best of the annotator's ability to diagnose the model's behavior by looking at the PDF, parsed XML, and extracted JSON. 

In any given paper, different values will be reported in different forms, such as yield strengths in table and testing temperature in narrative text. Our annotators were instructed to note the format of the majority of the ``key information'' for each paper, and based their failure evaluations based on that information. 

\subsection{Data format}

Figure \ref{fig:data_info_err_distribution} summarizes the data format of key information in both datasets, including both original rows in the dataset and rows the model failed to extract. Possible data formats include table, figures, narrative text, calculated, meaning manually calculated by the creator of the dataset from other information within the paper, and other, a miscellaneous category. 

Figure \ref{fig:data_info_err_distribution}A shows that, while a plurality of relevant MPEA data is located in tables, a significant percentage of the data is located in figures, which is inaccessible to any text-only extraction method. A significant percentage of relevant information is also located in written narrative form. Looking at the distribution of data format for rows the LLM failed to extract, we can see that it trivially misses 100\% of figure-formatted data, 2/3 of table-formatted data, and a relatively smaller 1/3 of narratively-described data.  Figure \ref{fig:data_info_err_distribution}B shows the same numbers for the diffusion dataset, where the vast majority of relevant data was reported in tables, which the LLM in almost every case failed to extract. 


\begin{figure*}[h!]
   

    \begin{subfigure}{.24\textwidth}
        \centering
        \includegraphics[width=1\linewidth]{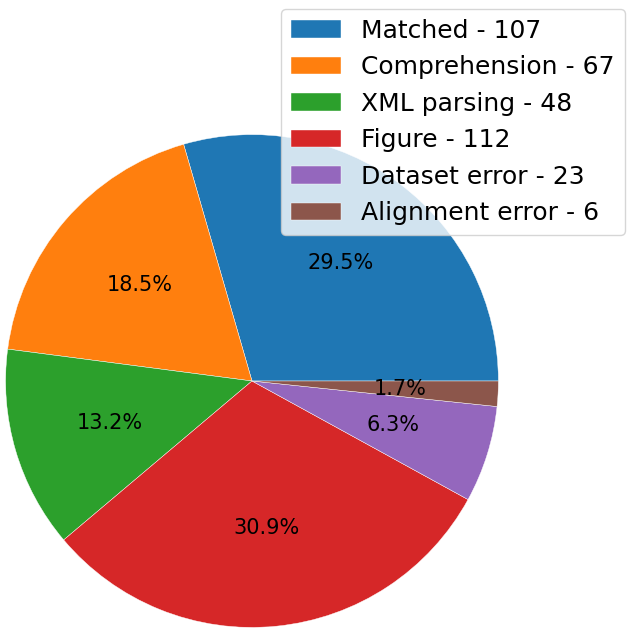} 
        \caption{MPEA misses/matches}
        \label{fig:sub-first}
    \end{subfigure}
        \begin{subfigure}{.24\textwidth}
        \centering
        \includegraphics[width=1\linewidth]{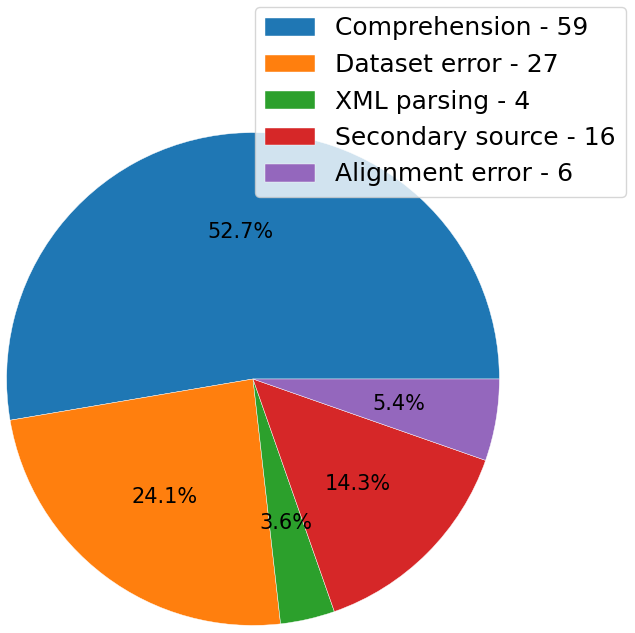}
        \caption{MPEA hallucinations}
        \label{fig:sub-first}
    \end{subfigure}
    \begin{subfigure}{.24\textwidth}
        \centering
          \includegraphics[width=1\linewidth]{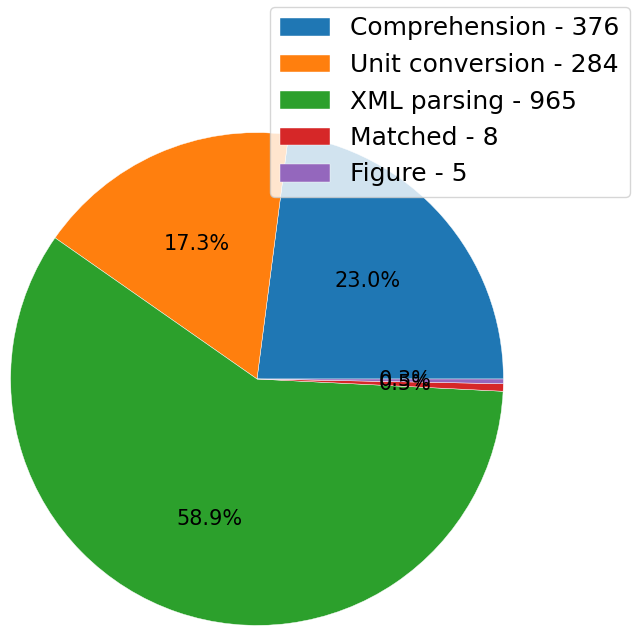} 
        \caption{Diffusion misses/matches}
        \label{fig:sub-second}
    \end{subfigure}
        \begin{subfigure}{.24\textwidth}
        \centering
        \includegraphics[width=1\linewidth]{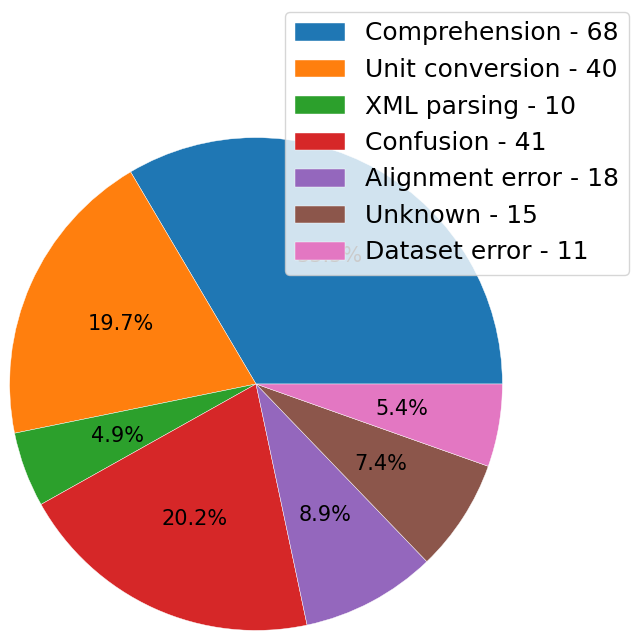}
        \caption{Diffusion hallucinations}
        \label{fig:sub-second}
    \end{subfigure}

    \caption{Proportions of differing error reasons, divided by error type (missed row vs. hallucination) and dataset. }
    \label{fig:error_reason_plot}
   
 \end{figure*}
\subsection{Error analysis}
\label{subsec:error_analysis}

Figure 4 summarizes annotated causes of error between the two datasets. It includes a separate plot for misses and matches (covering all the original rows in the dataset covered by the annotator), and one for hallucinations, indicating extracted rows that could not be matched to an original dataset row. Reported error types include \textbf{XML parsing errors}, where GROBID failed to extract key information from the PDF; \textbf{figure errors} where key information was stored in figures invisible to the model; \textbf{dataset errors} where the original dataset had missing or incorrect values which the model successfully extracted; \textbf{generalized comprehension errors} where the LM overlooked key information; \textbf{unit conversion errors} where values were reported in a different unit from that recorded in the dataset; \textbf{confusion errors} where the model confused one value for another; \textbf{secondary source}, where the model picked up values from other sources referred to in passing; and \textbf{alignment errors} where the alignment algorithm described above failed to match extracted to original rows. 

\paragraph{MPEA dataset.} 


Figure \ref{fig:error_reason_plot}A summarizes error causes reasons for MPEA original dataset rows (both matched and missed).  For the 368 ground-truth entries evaluated from the MPEA dataset, a total of 107 were matched (29.5\%). Of the missed 70.5\%, major sources of error were model comprehension problems (18.5\%), key information being presented in figures (30.9\%), XML parsing problems (13.2\%), and errors in the original dataset (6\%).  

Thus for this dataset the model's performance on the extraction task was hindered more by the form of data presentation than by limitations in comprehension ability.  The majority of annotated entries in the MPEA set that were missed by the model (160 of 256, or 62.5\%) were entries that were located in either a figure or a table that did not parse properly into the XML input.  For both of these cases, it is unreasonable to expect the extraction of information that was never presented to the model in the first place, so even though these database values could not be recreated, the failure does not provide much insight from a text comprehension perspective.  For the 67 missed MPEA entries that were comprehension-related (annotators manually determined that relevant information was present in text or a table that properly parsed into XML), 2/3 of them were values that appeared only in tables.  Cases of missed values in a table often occurred with tables that were relatively complex in structure, such as in  \citet{senkov2020temperature}, where information on yield strength is merged in with information on flow stress and categorized by temperature and relative contribution from material constituent structure--making it difficult to directly extract the yield strength as demanded by the dataset schema.

For values missed in the text itself, some of these errors can be attributed to cases where the value was mentioned in discussion around specific figure or table, making the text itself more difficult to interpret.  For example, in \citet{hu2019microstructure}, the discussion of MPEA yield strength in the text occurs in the excerpt ''[t]he maximum compressive strength is... for the NbZrTi and NbHfZrTi alloys, respectively," which is stated as part of a discussion around a related figure showing that this ``maximum compressive strength" corresponds to the yield strength. However, without the associated figure, the text alone is difficult to properly interpret.  Still, there were apparent cases of comprehension failure, such as in \cite{tseng2018effects}, where, for one of their target test temperatures, the authors plainly state ``the compressive yield strength was 1005 MPa,'' a point which the model failed to detect.   

Figure \ref{fig:error_reason_plot}B summarizes reasons for hallucinated rows in the MPEA dataset. The most common form of hallucination on the MPEA dataset were generalized miscomprehensions (59 cases, 52.7\%), often cases where the model identified a valid composition but none of the other relevant properties. The next most common form of reported hallucination (27, 25.6\%) were not actually hallucinations, but rather the model finding accurate compositions and property values that were not collected in the construction of the original dataset. The third most common type of ``hallucination'' were cases where composition/yield strength combinations reported in prior work were mentioned in the paper at hand, and then extracted by the model. XML parsing was not a major source of error for the MPEA dataset, where most of the included papers are relatively recent. 

\paragraph{Diffusion dataset} 

Figure \ref{fig:error_reason_plot}C summarizes the percentage of expert-annotated error reasons for diffusion dataset rows (both matched and missed).

The overall performance of the model was distinctly different for this dataset.  Out of a total of 1714 dataset entries, only 23 entries were able to be automatically aligned with original records, and 15 of these were missing so many other values as to be effectively useless. Thus, extraction almost completely failed for this model. By far the largest source of error was XML parsing issues, which were responsible for 965 misses. Errors in XML parsing were often related to the age of the publication.  Many of the older diffusion publications in the set have outdated typesetting practices and and non-standard table structures with multiply-merged cells, both of which created numerous issues in the model input information.

After XML issues, the next largest error source was generalized model comprehension, which was responsible for 387 (24.2\%) of misses. The model had significant difficulty correctly reporting textual values like melt descriptions, as these often required domain knowledge connections that were difficult to make in the context of the prompt.  For example, \citet{alletti2007halogen} studied diffusion in a Hawaiitic melt, but during database compilation, this `Hawaiitic' description was instead reported as `basalt' as part of a standardizing effort by \citet{zhang_diffusion_2010} to reduce the number of unique mineral terms in the database.  When the model retrieved summarizing values from the abstract of \citet{alletti2007halogen}, the melt was labeled as ``Hawaiitic melt from Mt. Etna", which is factually correct, but does not align with the database description of basalt. Other key forms of miscomprehension included misreading complex table formats like in the MPEA dataset, and erroneously preferring narratively-presented information to tabular information. In processing \cite{mackenzie_volatile_2008}, for instance, the model seems to have extracted values presented in the abstract rather than in the centerpiece tables. 

Finally, unit mismatches were a significant source of error, which were related to 17.1\% of all rows.  When the database was compiled by \citet{zhang_diffusion_2010}, the standard units of $m^2/s$ were chosen for the diffusion coefficient, but it is not uncommon for this coefficient to also be reported in $cm^2/s$, $\mu m^2/s$, or even as a logarithm ($log(D)$), depending on the context of the investigation.  Unit-based comprehension issues often involved failure to report diffusion coefficient values in the correct units, and failure to correctly convert logarithm-form diffusion coefficients. Without access to an external conversion tool, it is unreasonable to expect the LLM to perform these unit conversions. 

Figure \ref{fig:error_reason_plot}D summarizes hallucinated rows for the diffusion dataset.  Of the 203 reported hallucinations reported on the annotated portion of the Diffusion dataset, 109 (53.6\%) were annotated as comprehension issues, including ``confusion'' instances where the model confused the desired value for something else. An example is \cite{roselieb_tracer_2002}, where it incorrectly took Do values from the abstract. Another reported type hallucination was the retrieval of simplified or summarizing values.  The remaining hallucinations were a combination of XML issues, assignment comprehension issues, and entries not collected during dataset construction.

\section{Discussion}

The manual error analysis identifies several major barriers to effective ad-hoc schema-based information extraction from scientific articles. 

\paragraph{PDF parsing.} PDF parsing is a problem~\cite{gururaja2024collage} for older PDFs which, for some areas such as geology and certain subfields of materials science, can still contain valuable information worth extracting. This suggest a need for native LM support for PDFs, which, while present in major models such as Claude and ChatGPT via their interfaces, is not yet available in programmatic API form at the time of writing.

\paragraph{Figure comprehension.} The MPEA dataset demonstrates that a considerable amount of valuable quantitative scientific information is presented visually rather than in tabular form, suggesting a need for vision-language models capable of reading values from plots and figures. For tasks like determining yield strength from a load curve, a vision model would need to be able to recognize the curve itself and accurately read a singular specific point off of that curve, which may require specialized tuning. Recent work like \citet{shi_exploring_2023} provides a starting point for this.  

\paragraph{Unit conversion.} A characteristic problem in the diffusion dataset is the presentation of results in units like $log(m^2/s)$ or $cm^2/s$ when the schema requires $m^2/s$. A LM capable of normalizing across these different units would need access to external tools capable of doing so, like ToolFormer \cite{schick_toolformer_2023}.

\paragraph{Table comprehension.} By far the most common format for experimental results is tables, and table miscomprehensions, especially those associated with complex or non-standard table formats, formed the bulk of true comprehension errors made by the model on both datasets. Thus, a key research direction needs to be improved table comprehension specifically. 

\paragraph{Narrative context for tabular information.} While tabular data is typically (though not always) the centerpiece of a given paper, narrative information provides important context for understanding it. Constant values such as the temperature at which experiments were run is often presented narratively (which we want to extract), and selected values from a given table are often summarized narratively (which we don't want to extract). Secondarily-reported values from prior work also falls into this category, as it is often presented as context for the primary set of results. Thus, another important research direction toward ad-hoc scientific information extraction is to correctly parse the relationship between tabular and narratively-presented data, perhaps by explicitly treating tabular data as higher priority than narrative data when performing holistic extraction. 


\paragraph{Deeper schemas.} There are also significant challenges associated with the use of scientific terminology across different fields. This is particularly apparent for the diffusion dataset, which represents a significant effort by an experienced geologist to compile silicate diffusion information and present it in a form that translates across disciplines. The result of this effort shows through particularly in the melt and experiment descriptions in the compiled dataset, both of which have been standardized as much as possible. For an IE model to accurately recreate these types of standardized descriptions, it would need to understand the full schema associated with the standardization, and also be sufficiently flexible to generate the unique tokens associated with these schema labels, even when those tokens do not occur within the text itself. Creating a role and prompt for the model that permits this flexibility while also mitigating hallucination is a nontrivial task.







\section{Conclusion}

We perform an analysis of whether a representative contemporary large language model, GPT-4, is capable of performing complex ad-hoc scientific information extraction sufficient for replicating two materials property datasets. A detailed manual expert error analysis shows that, while the model shows promise for narrative and tabular data, there are significant barriers in both modality and comprehension ability that prevent the model from being adequate to the task using a baseline one-shot prompting approach. Nevertheless, our initial results on this only-recently-recognized information extraction paradigm and our careful analysis will help guide further work in this area. 


\section*{Limitations}


Our goal in this paper is to study the limitations of a representative contemporary large language model (GPT-4) in performing ad-hoc schema-based information extraction. Our analysis approach has its own limitations, however, including the assumption of representativeness in the one model we evaluate and the two materials datasets we evaluate it on. GPT-4's behavior may not generalize to other models, nor do other manually-curated scientific datasets necessarily conform to the structure of the two that we study. There will need to be extensive further experimentation with a wider variety of model and datasets to fully characterize the research frontier for this topic. 

\section*{Acknowledgments} This work was supported by the OpenAI Researcher Access Program.


%


\bibliography{matsci_ie_refs,materials_lit}
\bibliographystyle{acl_natbib}

\appendix

\section{Appendix}
\label{sec:appendix}

\subsection{Dataset column descriptions}

Table \ref{tab:dataset_column_descriptions} describes each major column in each dataset. 

\begin{table*}[]
\scriptsize
\centering
\begin{tabular}{@{}llll@{}}
\toprule
\multicolumn{2}{c}{\textbf{MPEA dataset}}                                                                                                        & \multicolumn{2}{c}{\textbf{Diffusion dataset}}                                                                                                \\ \midrule
\textbf{Column}                        & \textbf{Definition}                                                                                     & \textbf{Column}    & \textbf{Definition}                                                                                                      \\ \midrule
FORMULA                                & Composition of the alloy                                                                                & melt               & \begin{tabular}[c]{@{}l@{}}Common name of substance being \\ diffused into (e.g. Andesite, Basalt11)\end{tabular}        \\
PROPERTY: Microstructure               & Crystal structure of alloy                                                                              & diffusing species  & Element diffusing into melt                                                                                              \\
PROPERTY: Processing method            & \begin{tabular}[c]{@{}l@{}}Synthesis method (e.g. CAST, ANNEAL, \\ POWDER, WROUGHT, OTHER)\end{tabular} & type of diffusion  & \begin{tabular}[c]{@{}l@{}}How atoms are diffusing into melt \\ (e.g. Tracer, SEBD, FEBD, self)\end{tabular}             \\
PROPERTY: BCC/FCC/other                & Categorical subset of crystal structure                                                                 & type of experiment & Physical experiment setup                                                                                                \\
PROPERTY: grain size ($\mu$m)          & Grain size within alloy                                                                                 & T(K)               & Experiment temperature                                                                                                   \\
PROPERTY: Exp. Density (g/cm$^3$)      & Experimental density of alloy                                                                           & P(MPa)             & Experiment pressure                                                                                                      \\
PROPERTY: Computed Density (g/cm$^3$)  & Computed density of alloy                                                                               & D(m2/s)            & Diffusivity; speed of diffusion                                                                                          \\
PROPERTY: HV                           & Vicker's hardness of alloy                                                                              & 1s error           & \cellcolor[HTML]{FFFFFF}{\color[HTML]{222222} SD error in diffusivity when reported.}                                    \\
PROPERTY: Type of test                 & Type of test (C=compression, T=tensile)                                                                 & SiO2               &                                                                                                                          \\
PROPERTY: Test temperature ($^\circ$C) & Testing temperature                                                                                     & TiO2               &                                                                                                                          \\
PROPERTY: YS (MPa)                     & Yield strength                                                                                          & Al2O3              &                                                                                                                          \\
PROPERTY: UTS (MPa)                    & Ultimate tensile strength                                                                               & FeOt               &                                                                                                                          \\
PROPERTY: Elongation (\%)              & Elastic elongation at break                                                                             & MnO                &                                                                                                                          \\
PROPERTY: Elongation plastic (\%)      & Plastic elongation at break                                                                             & MgO                &                                                                                                                          \\
PROPERTY: Exp. Young modulus (GPa)     & Experimental Young's Modulus                                                                            & CaO                &                                                                                                                          \\
PROPERTY: Computed Young modulus (GPa) & Computed Young's modulus                                                                                & Na2O               &                                                                                                                          \\
PROPERTY: O content (wppm)             & Oxgyen content                                                                                          & K2O                &                                                                                                                          \\
PROPERTY: N content (wppm)             & Nitrogen content                                                                                        & P2O5               &                                                                                                                          \\
PROPERTY: C content (wppm)             & Carbon content                                                                                          & H2Ot               & \multirow{-11}{*}{\begin{tabular}[c]{@{}l@{}}Percentage of each compound in the \\ composition of the melt\end{tabular}} \\
                                       &                                                                                                         & Composition unit   & Unit of compound composition                                                                                             \\
                                       &                                                                                                         & dry total          & Non-H2O fraction of melt                                                                                                 \\ \bottomrule
\end{tabular}
\caption{Descriptions of individual columns in MPEA and diffusion datasets.}
\label{tab:dataset_column_descriptions}
\end{table*}

\begin{figure*}
    \centering
    \includegraphics[width=1\linewidth]{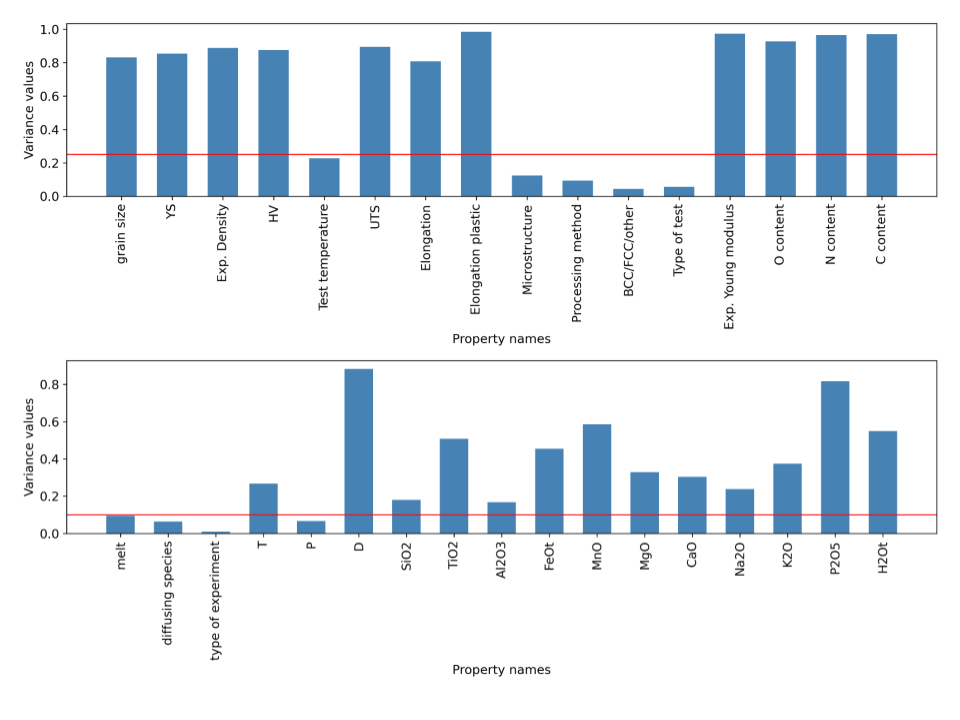}
    \caption{Plot to visualize the variance of properties found in two datasets. High variance means properties that are altered in experiments and low variance show experimental constants.}
    \label{fig:property_variance_visualization}
\end{figure*}

\subsection{Full prompt}
\label{sec:full_prompt_details}
The prompts used for both extraction tasks are given below.
\subsubsection{Role setting for both tasks}
I am a helpful assistant capable of extracting information from text. I will not generate any new tokens, only extract tokens containing information from the text you provide. If provided with a schema, then I will follow it and return the information in format of the schema.

\subsubsection{Prompt for MPEA extraction}
Extract all information relating to every High Entropy Alloys (HEA) from the following text: \{exemplar text / other text to extract from\} using the following schema: {schema}. Return a list of schemas in JSON format for every unique compound.

Additional context: HEA composition sometimes has variable ratio denoted by 'x', if so, then replace x with a float or integer as applicable by the information in the paper.

Restriction: Do not extract any tokens if a HEA material does not have a property mentioned in the schema. If a property is not available then return 'No information'. Do not violate the schema.

\subsubsection{Schema for MPEA extraction}
\begin{small}
\begin{verbatim}
{   
    "high entropy alloy formula": {"type": "string"},
    "microstructure": {"type": "string"},
    "processing method": {"type": "string"},
    "BCC/FCC/other": {"type": "string"},
    "grain size": {"type": "float"},
    "experimental density": {"type": "float"},
    "hardness": {"type": "float"},
    "type of test": {"type": "string"},
    "test temperature": {"type": "float"},
    "yield strength": {"type": "float"},
    "ultimate tensile strength": {"type": "float"},
    "elongation": {"type": "float"},
    "elongation plastic": {"type": "float"},
    "experimental young modulus": {"type": "float"},
    "oxygen content": {"type": "float"},
    "nitrogen content": {"type": "float"},
    "carbon content": {"type": "float"}
}
\end{verbatim}
\end{small}
\subsubsection{Exemplar output for MPEA extraction}

One entry from the one-shot output exemplar that we used is given below. The actual output contains 5 of these entries in a JString list.\newline

\begin{small}
    \begin{verbatim}
{
    "high entropy alloy formula": 'NbMoTaWVCr',
    "microstructure":'BCC+Laves+Sec.',
    "processing method":'POWDER',
    "BCC\/FCC\/other":'other',
    "grain size": 0.54,
    "experimental density": 'No information',
    "hardness": 1072.0,
    "type of test":'C',
    "test temperature": 25.0,
    "yield strength": 'No information',
    "ultimate tensile strength": 'No information',
    "elongation": 'No information',
    "elongation plastic": 'No information',
    "experimental young modulus": 'No information',
    "oxygen content": 7946.0,
    "nitrogen content": 'No information',
    "carbon content": 'No information'
}
    \end{verbatim}
\end{small}

\subsubsection{Prompt for Diffusion extraction}
Extract all information relating to diffusion of elements into a silicate melt from the following text: \{one shot text / other text to extract from\} using the following schema: \{schema\}. Return a list of schemas in JSON format for every unique compound.

 Additional context: there will generally be one or multiple liquids, and one or more elements that move through that liquid at a certain speed. That speed is called the “diffusivity” or the ``diffusion coefficient'' D(m\^2/s). Each element will have a unique diffusivity in each melt and temperature and pressure, so it is important to keep track of every combination.
 
Restriction: Do not extract any tokens if no relevant property is present as mentioned in the schema. If a property is not available then return 'No information'. Do not violate the schema.

\subsubsection{Schema for Diffusion extraction}

\begin{small}
    \begin{verbatim}
{
    'melt': {'type': 'string'},
    'diffusing species': {'type': 'string'},
    'type of experiment': {'type': 'string'},
    'test temperature': {'type': 'float'},
    'pressure': {'type': 'float'},
    'diffusivity': {'type': 'float'},
    'SiO2': {'type': 'float'},
    'TiO2': {'type': 'float'},
    'Al2O3': {'type': 'float'},
    'FeOt': {'type': 'float'},
    'MnO': {'type': 'float'},
    'MgO': {'type': 'float'},
    'CaO': {'type': 'float'},
    'Na2O': {'type': 'float'},
    'K2O': {'type': 'float'},
    'P2O5': {'type': 'float'},
    'H2Ot': {'type': 'float'}
}

    \end{verbatim}
\end{small}

\subsubsection{Exemplar output for Diffusion extraction}
The following is one of 21 entries in the one-shot exemplar output. The actual output is a list of such entries given to the model as a JString.
\begin{small}
    \begin{verbatim}
{
    "melt": "NCMAS6",
    "diffusing species": "Fe",
    "type of experiment": "electrochemistry",
    "test temperature": 1573.15,
    "pressure": "No information",
    "diffusivity": 1.35e-07,
    "SiO2": 80.6793201360426,
    "TiO2": "No information",
    "Al2O3": 0.0,
    "FeOt": "No information",
    "MnO": "No information",
    "MgO": 0.0,
    "CaO": 14.11921335907197,
    "Na2O": 5.201466504885413,
    "K2O": "No information",
    "P2O5": "No information",
    "H2Ot": "No information"
}        
    \end{verbatim}
\end{small}

\subsubsection{Exemplar text for both tasks}
The text is too long to be presented here but can be viewed in~\cite{long2019fine} (MPEA) and \cite{russel2004effect} (Diffusion). Input text comprised of the abstract along with other sections and tables (including their captions) sequentially as present in the paper.

\subsection{Illustrative examples of errors}

As mentioned in the discussion section we found errors occurring primarily due to some reasons like table comprehension, text comprehension, ground-truth information present in plots/images, unit conversion problems, poor XML parsing, simplification of information, and partial information extraction from secondary sources. We wanted to show what those problems look like so in the subsections below we will give an example for each of the errors. We also show an example of an instance when the model was actually correct, in-spite not being matched with the human annotator.

\subsubsection{Table comprehension}
\begin{figure}[!h]
    \centering
    \includegraphics[width=\linewidth]{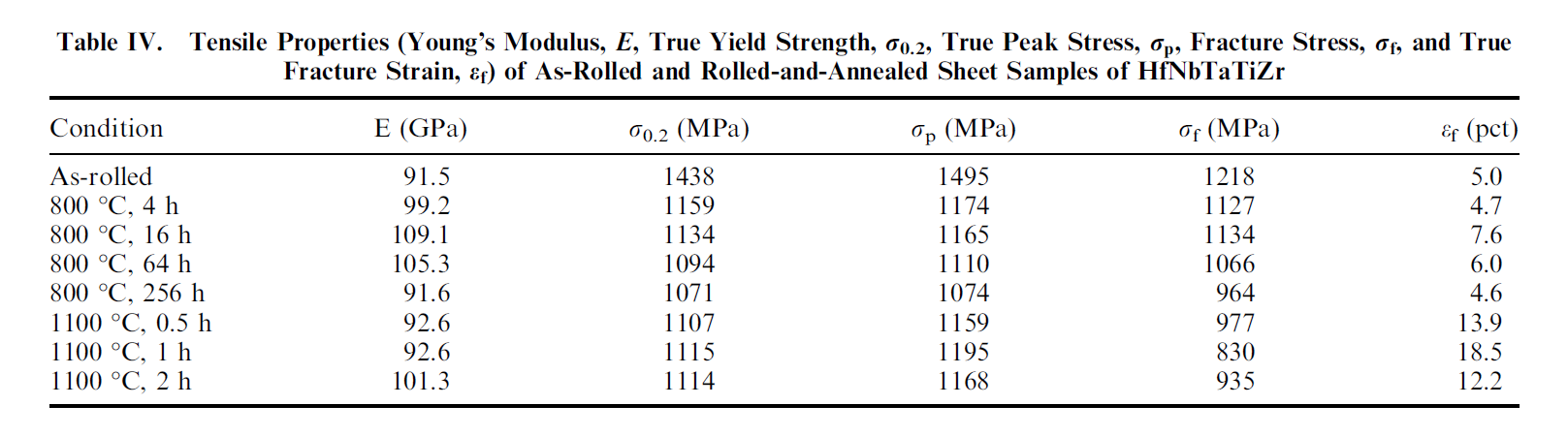}
    \caption{This is an example table that the model failed to comprehend even when the XML parsing was perfect.}
    \label{fig:table_comprehension}
\end{figure}

In Figure~\ref{fig:table_comprehension} taken from~\citet{senkov2018effect} we found that the model was able to extract the yield strength of three out of eight yield strength value even when the caption clearly mentioned which column corresponds to the true yield strength.

\subsubsection{Text comprehension}
\begin{figure}[!h]
    \centering
    \includegraphics[width=1\linewidth]{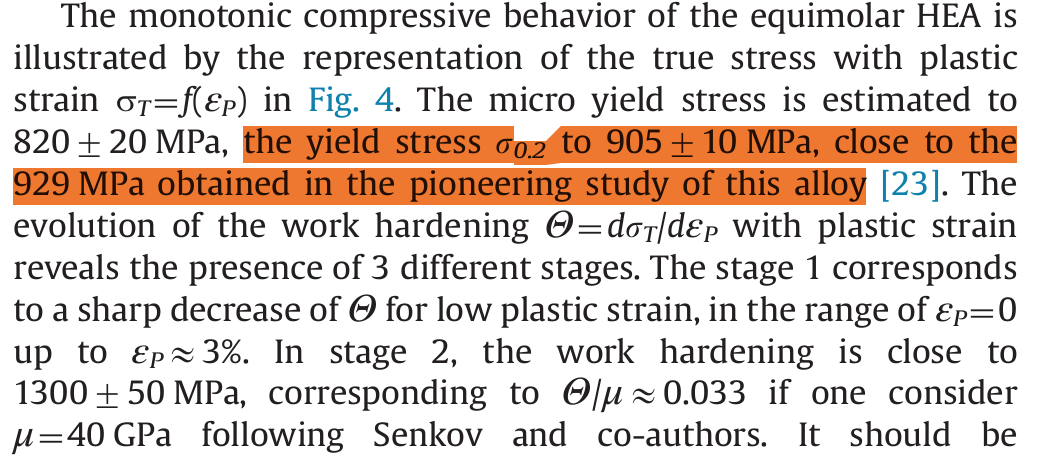}
    \caption{This image shows the paragraph which the model failed to comprehend and did not extract the yield strength value (highlighted text).}
    \label{fig:text_comprehension}
\end{figure}

The text in Figure~\ref{fig:text_comprehension} is a paragraph of text taken from~\citet{couzinie2015room}, where the model could not understand that the text is mentioning information about yield strength.

\subsubsection{Information in plots/images}
\begin{figure}
    \centering
    \includegraphics[width=1\linewidth]{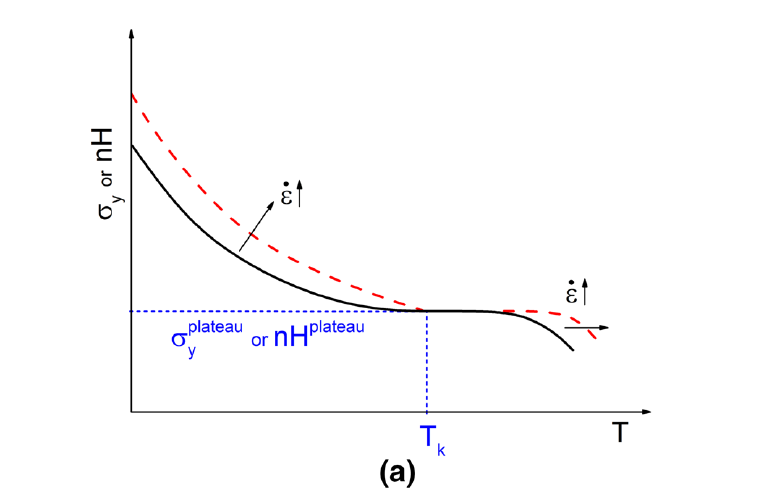}
    \caption{Figure of a plot that contains information about the relationship between yield strength and temperature.}
    \label{fig:info_in_plot}
\end{figure}

A lot of missed extraction were due to the information being present in plots. As the XML parsing cannot process the image this information is skipped. In Figure~\ref{fig:info_in_plot}, we can see such a plot where the information about the yield strength was present in the plot~\cite{chen2018contribution}.

\subsubsection{Unit conversion}
\begin{figure}[h!]
    \centering
    \includegraphics[width=1\linewidth]{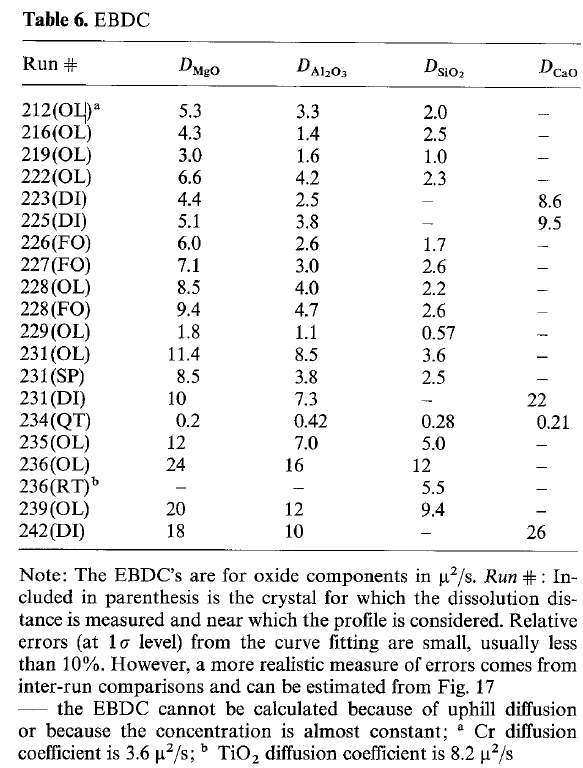}
    \caption{Example of a table with failed extraction due to unit conversion error. The table presented rate of diffusion in $\mu / s^2$ whereas the gold extraction where in $m / s^2$.}
    \label{fig:unit_conversion_error}
\end{figure}

In multiple situation of the diffusion dataset we saw that the model extracted the correct information as present in the paper but as the units are not converted as the gold-extractions therefore we could not match them. An example this error can be found in Figure~\ref{fig:unit_conversion_error} from~\citet{zhang1989diffusive}.

\subsubsection{XML parsing}
\begin{figure} [h!]
    \centering
    \includegraphics[width=1\linewidth]{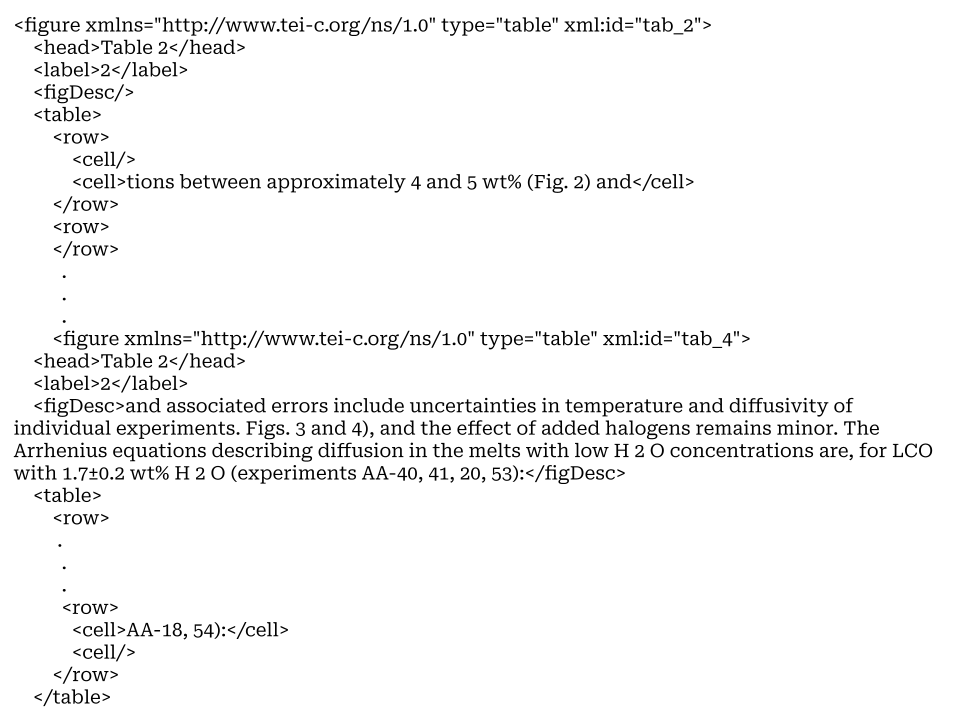}
    \caption{The given XML text was the parsed text from the PDF file.}
    \label{fig:xml_parsing}
\end{figure}

\begin{figure}[h!]
    \centering
    \includegraphics[width=1\linewidth]{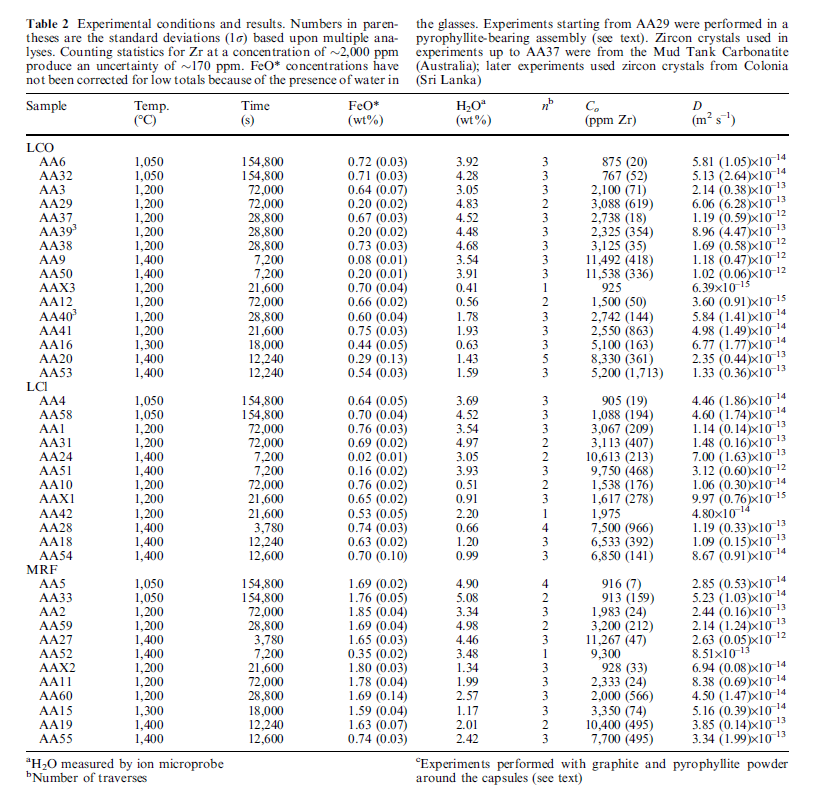}
    \caption{Table that was parsed incorrectly leading to information loss.}
    \label{fig:xml_parsing_table}
\end{figure}

XML parsing of some older papers from the diffusion dataset was not of good quality and this contributed to some failed extraction. One such example can be found in Figure~\ref{fig:xml_parsing} from~\citet{baker2002effect}. The XML created from the PDF file was not parsed properly and the actual values in the cells were just missed during the parsing process. The actual table is provided in Figure~\ref{fig:xml_parsing_table}.

\subsection{Simplification}

\begin{figure}[h!]
    \centering
    \includegraphics[width=1\linewidth]{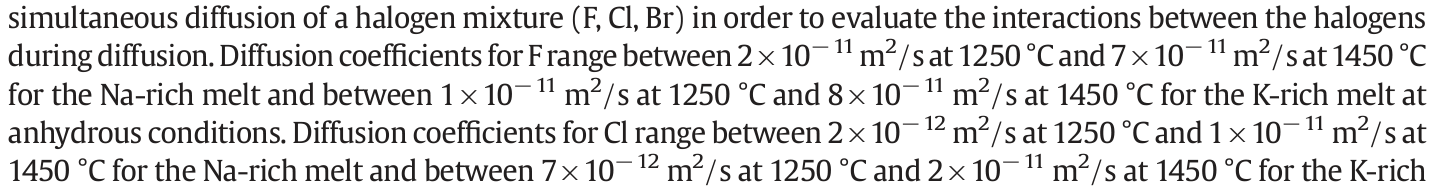}
    \caption{Example of text simplification performed by the model.}
    \label{fig:error_simplification}
\end{figure}

In Figure~\ref{fig:error_simplification}, we can see the abstract from \citet{balcone2009f}, that mentions different values of diffusion rates at different temperatures. The model extracted all the diffusion rates correctly but missed the varying temperatures associated with them and associated them with one temperature value, thereby simplifying the text. 

\subsection{Secondary source}
\begin{figure}[h!]
    \centering
    \includegraphics[width=1\linewidth]{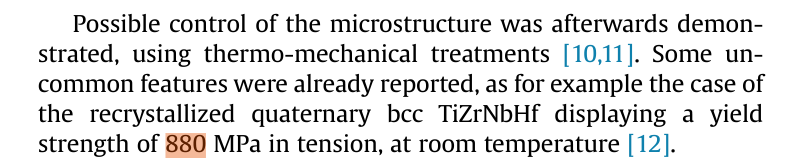}
    \caption{An example of secondary information that was collected by the model without a detailed context around it.}
    \label{fig:secondary_error}
\end{figure}

In \citet{lilensten2018study}, we can can find values of yield strength reported from other papers that were reviewed by the author. The authors reported these values as a summary of the entire paper without going into the details and that can be seen in Figure~\ref{fig:secondary_error}. We considered these extraction as errors as they are from secondary sources. In the given example the model extracted the value 880 MPa along with the generic formula of TiZrNbHf as mentioned in the text without any other information.

\subsection{Actually correct}
\begin{figure}[h!]
    \centering
    \includegraphics[width=1\linewidth]{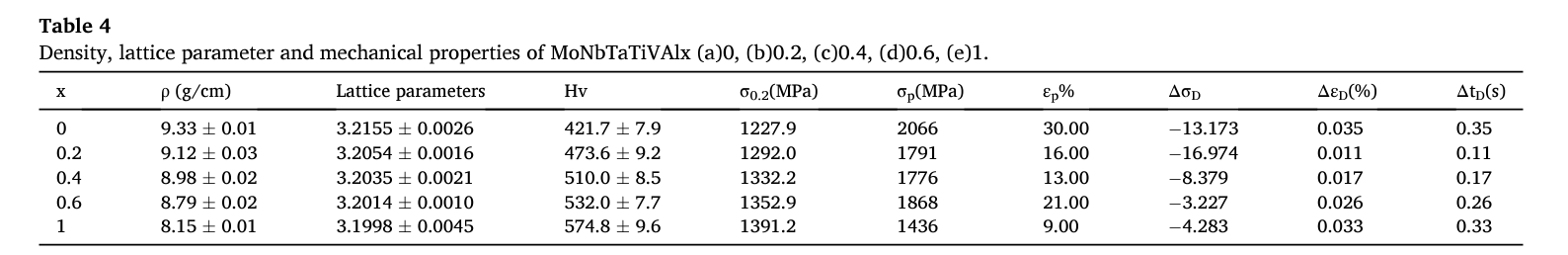}
    \caption{Model extracted values from this table as found in the paper.}
    \label{fig:actually_correct}
\end{figure}

In this example taken from \citet{ge2020effects} the model extracted correct yield strength value from table (Figure~\ref{fig:actually_correct}) and formula as reported. The gold extractions reported by the expert had rounded off yield strength values and the formula was reported as ratio of elements, leading to matching failure.

\end{document}